# Automated Clinical Coding: What, Why, and Where We Are?

Hang Dong[1,9], Matúš Falis[2], William Whiteley[3], Beatrice Alex[2,5], Joshua Matterson[6,7], Shaoxiong Ji[8], Jiaoyan Chen[9], Honghan Wu[4]

[1]Centre for Medical Informatics, Usher Institute, University of Edinburgh, Edinburgh, United Kingdom; [2]School of Informatics, University of Edinburgh, Edinburgh, United Kingdom; [3]Centre for Clinical Brain Sciences, University of Edinburgh, Edinburgh, United Kingdom; [4]Institute of Health Informatics, University College London, London, United Kingdom; [5]Edinburgh Futures Institute, Edinburgh, United Kingdom; [6]Epic Systems Corporation, Wisconsin, United States; [7]University College London Hospitals NHS Foundation Trust, Clinical Research Informatics Unit, London, United Kingdom; [8]Department of Computer Science, Aalto University, Espoo, Finland; [9]Department of Computer Science, University of Oxford, Oxford, United Kingdom

**Abstract**

*Clinical coding is the task of transforming medical information in a patient's health records into structured codes so that they can be used for statistical analysis. This is a cognitive and time-consuming task that follows a standard process in order to achieve a high level of consistency. Clinical coding could potentially be supported by an automated system to improve the efficiency and accuracy of the process. We introduce the idea of automated clinical coding and summarise its challenges from the perspective of Artificial Intelligence (AI) and Natural Language Processing (NLP), based on the literature, our project experience over the past two and half years (late 2019 - early 2022), and discussions with clinical coding experts in Scotland and the UK. Our research reveals the gaps between the current deep learning-based approach applied to clinical coding and the need for explainability and consistency in real-world practice. Knowledge-based methods that represent and reason the standard, explainable process of a task may need to be incorporated into deep learning-based methods for clinical coding. Automated clinical coding is a promising task for AI, despite the technical and organisational challenges. Coders are needed to be involved in the development process. There is much to achieve to develop and deploy an AI-based automated system to support coding in the next five years and beyond.*

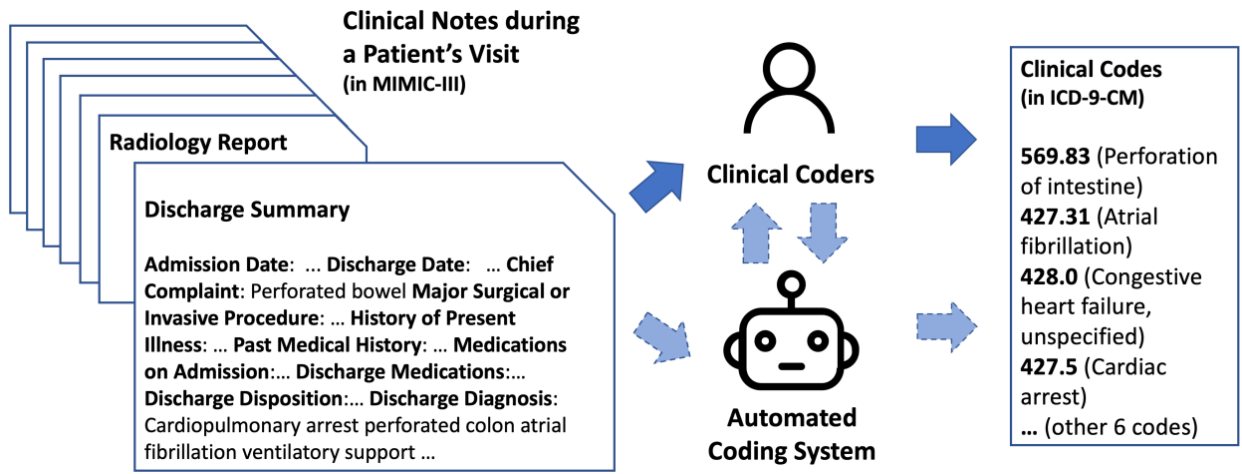

**Figure 1. An example of clinical coding, manual and automated** (linked with solid and dashed arrows, respectively), with ICD-9-CM codes from a clinical note in the MIMIC-III dataset [15] of ICU patients in 2001-2012 in a hospital in the US. Dashed arrows between clinical coders and the automated coding system suggest potential interactions between them, while this is yet to be considered in many clinical coding systems. Note that the format of data and clinical codes does not reflect the situation of other regions in the world - for example, in the UK, where data may be less structured and there is no universal discharge summary format available.

**Main Text**

**Introduction: what is (automated) clinical coding?**

**Clinical coding** is the task of transforming medical records, usually presented as free texts written by clinicians, into structured codes in a classification system like ICD-10 (International Classification of Diseases, Tenth Revision). For example, in Scotland, this means to apply a standard process to classify information about patients into appropriate diagnosis and procedure codes in ICD and OPCS (OPCS Classification of Interventions and Procedures), finally contributing to the Scottish Morbidity Records (SMR01) national dataset [1]. The purpose of clinical coding is to provide consistent and comparable clinical information across units of care and over time. The resulting national data are used to support areas, such as health improvement, inform healthcare planning and policy and add to the epidemiological understanding of a wide variety of conditions, so confidence in the data is essential. Also, codes are mainly used for billing purposes in the US [2]. For introductory slides about clinical coding in the UK provided by NHS Digital, see *Clinical coding for non coders* [3].

Clinical coding is a non-trivial task for humans. The process of coding usually includes data abstraction or summarisation [4]. More specifically, an expert clinical coder is expected to decipher a large number of documents about a patient's episode of care, and to select the most accurate codes from a large classification system (or an ontology), according to the contexts in the various documents and the regularly updated coding guidelines. For example, coding in the US adopts the International Classification of Diseases, Tenth Revision, Clinical Modification (ICD-10-CM), which has around 68,000 diagnosis codes [5]; ICD-10 is also the main classification for coding in the UK. There is a standard process for manual coding to ensure data consistency: textual analysis, summarisation, and clearly defined steps to classification into codes (or the four steps of *analyse, locate, assign, and verify* as suggested by the NHS digital in the coding standard of 2021 [6, p.11]). The process minimises the risk of introducing variations caused by artefacts (potentially leading to wrong decision making), thus collecting and analysing data and applying the standard is important. There are regularly updated guidelines and standards for coding (e.g., in Public Health Scotland [7]). Usually, it can take months or longer to train an expert clinical coder in the NHS (National Health Service) in the UK, and worldwide [8].

**Automated clinical coding** is the idea that clinical coding may be automated by computers using AI techniques, e.g., NLP and machine learning [9]. It is a branch of computer-assisted coding (CAC) [10]. In recent years, AI has been considered a promising approach to transforming healthcare by intelligently processing the increasing amount of data with machine learning and NLP techniques [11]. Automated clinical coding is a potential AI application to facilitate the administration and management of clinical records in the hospital and medical research. There has been a surge of articles for automated clinical coding with deep learning (as the current mainstream approach of AI) in the last few years, as reviewed in recent surveys [12-14].

However, while there is some progress for automated clinical coding, the task is far from solved. For the last two years and more, we have been working on the task and discussing it with practitioners of clinical coding and clinicians from Scotland and the UK. We illustrate the manual and automated clinical coding process, and their potential interactions, in Figure 1. In this paper, we aim to summarise the technical challenges of clinical coding, mainly related to deep learning, and propose directions for future research in this area.

**Why do we need automated clinical coding?**

There are some major reasons that automated clinical coding can be helpful. First, **manual coding is time-consuming**. A clinical coder in NHS Scotland usually codes about 60 cases a day (equivalent to 7-8min for each case) and an NHS coding department of around 25 to 30 coders usually codes over 20,000 cases per month. Even so, there is usually a backlog of cases to be coded, which can take several months or more (e.g., over a year [16]). Second, **manual coding may be prone to errors**. This may be due to incompleteness in a patient's data, subjectivity in choosing the diagnosis codes, lack of coding expertise, or data entry errors [4]. The average accuracy of coding in the UK was around 83% with a large variance among studies (50-98%) [17]. In Scotland, the accuracy of coding is very high [18] (e.g., in the assessment during 2019-2020, achieved 92.5% for 3-digit code accuracy and 88.8% for 4-digit code accuracy of main conditions), yet still not perfect and under-coding occurs (for around 20% of the common conditions). On the other hand, **computer-assisted coding could improve the accuracy, quality, and efficiency of manual coding**, according to a recent, qualitative literature review [10]. We believe that with recent AI technologies (e.g., NLP), automated coding has the potential to better support clinical coders. We mostly focus on the case that AI directly contributes to assigning clinical codes.

**Why is automated coding a complex problem to solve?**

While humans can achieve high accuracy in clinical coding, the standard procedure, text analysis, text summarisation, and classification into codes, poses immense challenges for computer-based systems. This requires Natural Language Understanding (NLU), one of the classical but largely unsolved areas of AI [19-20], and the linking of natural language to knowledge representations like the ICD-10 classification system. Also, this clinical task poses more specific challenges compared to common NLU tasks. From our experience, these relate mainly to the following difficulties:

1. **Clinical documents are variously structured, notational, lengthy, and incomplete**. Clinical coding requires the understanding of texts in clinical documents, which is usually different from other types of documents like publications or texts from social media. They have variable document structures, they can be lengthy (on average around 1500 words [21] in only the discharge summaries in a US intensive care dataset, MIMIC-III [15]), and use terse abbreviations and symbols [9,22] (e.g., "a [xx] y/o M w/ Hep C, HTN, CKD, a/w HTN emergency" in a discharge summary and the use of "?" to denote uncertainty and "+" to denote a positive test in MIMIC-III). Coding also requires the understanding of the entirety of a patient's records, which includes multiple types of documents (e.g., discharge summaries, radiology reports, pathology reports, etc.). These documents are not always in a structured format and are sometimes incomplete or missing.

2. **Classification systems used for coding are complex and dynamic**. The ICD-10-CM system (implemented in the US in 2015) has around 68,000 diagnosis codes in a large hierarchy, 5 times more than the previous ICD-9-CM (used in MIMIC-III) [5]. The ICD-11 system [23] (or ICD-11-MMS, ICD-11 for Mortality and Morbidity Statistics, came into effect in early 2022, but is yet to be used in practice in the US or the UK at the time of writing) "contains around 17,000 unique codes for injuries, diseases and causes of death, underpinned by more than 120,000 codable terms" and can code "more than 1.6 million clinical situations" using code combinations [24]. ICD-11 also introduces significant changes in chapter structure, diagnostic categories, diagnostic criteria, etc., for example, in psychiatric classification [25]. ICD-11-MMS has a similar structure as in ICD-10 with more chapters, but distinct from previous versions, ICD-11-MMS has its backbone as a semantic network ("Foundation Component"), a large and deep polyhierarchy (i.e. children can have more than one parents) of medical concepts, where ICD-11-MMS is derived from; coding with ICD-11-MMS also allows "post-coordination" that uses code combinations to express complex phenotypes of a patient [26] and more details and examples are in the ICD-11 reference guide [27]. Besides, to support the localisation of ICD systems, classification standards are updated regularly (e.g., usually every few months in Public Health Scotland [7]). Automated clinical coding needs to work with dynamic and complex classification systems.

3. **The social-technical issues with automated clinical coding systems are still to be explored**. From the perspective of information systems, transitioning to a (semi-)automated coding environment in a national healthcare system is more challenging than the technical issues themselves. How do coders interact with an AI-based CAC system (as modelled in Figure 1)? How to present the information in an automated coding system so that coders will easily ignore errors and make the most use of the correct automatic codes? Will coders trust such a system? How will the role of coders change (e.g., from coders to coding editors or coding analysts)? What new skills will coders need? [10]

**How to solve automated clinical coding: symbolic or neural AI?**

The two main schools of thought of AI have been either a *symbolic, knowledge-based* approach or a *neural network* (which further developed into deep learning) based approach [19]. Putting them into the task of clinical coding, the symbolic AI approach aims at making the use of symbols and rules to represent and model the standard practice that clinical coders apply in their work. The neural network and deep learning approach aims at learning a complex function to match a patient's information to the appropriate set of medical codes. This function is learned from the training data. From the historical perspective, symbolic AI, as the mainstream approach from 1950 to the early 1980s, did not scale up to complex real-world scenarios, for example, to model the natural language that people use in their daily life [19-20]. Neural networks returned in the mid-1980s with *machine learning* in general. *Deep learning* methods became the mainstream of AI after 2011 [20], continuing to evolve today [28].

Coming back to automated clinical coding, while the task has been studied for around 50 years (with the earliest studies around 1970 [29]), the current deep learning-based methods have a short history. Prior to deep learning, most studies use rules (regular expressions, logic expressions, and keywords) with feature engineering methods for text classification [9,14]. The issue with pure rule-based methods is that it is not straightforward and it can be time-consuming to extend rules to tens of thousands of codes and their varieties, and inter-relations among codes; this thus needs the support of machine learning with textual features for classification, and historically, some of the classifiers were Decision Trees, Support Vector Machine (SVM), etc. [9,14,30]. Still, rule-based methods like using regular expressions to match various textual descriptions can result in high precision in coding (yet low recall), and have been used to support human coding to largely improve coding efficiency [31].

Only since around 2017 [32-33], deep learning has been applied to automated coding and there are abundant studies in this area (reflected in recent surveys [12-14] and curation of papers in automated medical coding [34]). Unlike rule-based and traditional machine learning methods, pure deep learning methods do not require expert rules and hand-crafted textual features, thus easily applicable, while achieving better overall performance by learning from a sufficient amount of data [33]. Most of the studies formulate the task as a multi-label classification problem [35], while some studies formulate the task as a concept extraction or a Named Entity Recognition and Linking (NER+L) problem [36-37]. Though it seems that deep learning is the main method applied to automated clinical coding, we argue that there is still an important need for knowledge-based approaches in this area, and a better solution is to combine both schools of thought in the design of an automated clinical coding system. A recent trend is *knowledge-augmented deep learning* methods, where several studies used various *embedding*-based approaches to incorporate knowledge graphs into deep learning (to name a few [38-41]) or directly integrated the subsumption relations of codes into the model [42] and the evaluation [43]; however, the knowledge used is usually limited to the definition and hierarchies in the target ontology ICD-9 (except Freebase in Teng *et al.*, 2020 [39]), while the other vast number of clinical ontologies (e.g., UMLS, SNOMED-CT, and others) are not leveraged with the multi-label classification approach; also other information in the ontologies like axioms, logical expressions, and class attributes have not been leveraged. Coding standard and guidelines have also not been leveraged to enhance deep learning, where a challenge would be the need to extract and represent the knowledge from them, which varies by locations and requires input from coding experts.

**How do state-of-the-art deep learning models work so far?**

*Coding tasks involving complex reasoning, such as those in which disparate pieces of information must be connected, are a difficult challenge for current NLP systems.* – Kukafka *et al.*, 2006 [44], and also quoted in Stanfill *et al.*, 2010 [9]

Clinical coding is a complex testbed for contemporary AI, especially for machine learning and deep learning applied to NLP. During the last few years, the problem itself elicits applied and theoretical research on text representation learning [21,45], multi-task learning [42,46], zero-shot learning [38,47], meta-learning [48], multi-modal learning [49], etc. The pursuit of a full-fledged deep learning-based clinical coding system, however, is far from being achieved: **at the time of writing, the best Micro-F1 score (a harmonic mean of precision and recall evaluated based on pairs of a patient's information and a code) on the full 8,932 ICD-9 codes for the MIMIC-III data was under 60% (between 58%-60%)** [46,50-53]. MIMIC-III discharge summaries [15], although coded with the older and obsolete version of ICD (ICD-9-CM, the ninth version, Clinical Modification), are the main dataset used for benchmarking [21]. This dataset is also now older (collected over ten years ago, from 2001-2012), and only represents an intensive care dataset in the US, thus not representative of the documents available in the UK or other regions.

The main principle of the current deep learning approach is to find a complex function (non-linear and constructed by multiple layers) to match a clinical note of a patient's visit to a set of codes. As we introduced earlier, this is the multi-label classification setting. This approach, however, has several major limitations when applied to clinical coding:

1. **Handling unseen, infrequent, and imbalanced labels**: In the MIMIC-III dataset, around 5,000 codes appear fewer than 10 times in the training data and over 50% of codes never appear [38]. Vanilla deep learning models rely on large amounts of data for training and fail completely for new or unseen labels. Multi-label classification is also very challenging, especially when there are many labels or when the labels are imbalanced.

2. **Lack of symbolic reasoning capabilities**: Manual coding involves reasoning beyond just locating concepts in the notes. The coders sometimes need to connect different pieces of information together [9,44]. The information from different sources may even be *contradictory* to each other for the same patient. Their decisions are based on a standard coding process, aided by coding guidelines [6]. Deep learning, on the other hand, tries to simply learn from the labelled data the association between texts and codes in different (pre-trained) embedding spaces, without explicitly modelling the reasoning process. Human-like reasoning may be supported by knowledge-based techniques, which can potentially boost the performance and explainability of coding of deep learning methods. The reasoning may include formalising coding guidelines into logical expressions [30] and creating regular expressions to capture various diagnosis descriptions of a code [31], and leveraging various semantics in knowledge graphs constructed from several linked ontologies including the target ICD hierarchy.

3. **Handling long documents**: Looking for the relevant information of a code from a long document poses a "needle-in-the-haystack" problem. The recent Transformer-based pre-trained language models (e.g., BERT, Bi-directional Encoding Representations from Transformers [54]) usually require a limited length of up to 512 sub-word tokens (where a word can be tokenised into several sub-words) as input due to the memory-demanding self-attention mechanism, while discharge summaries *alone* in MIMIC-III have on average around 1,500 tokens or words [21] and up to over 10,000 tokens, not counting other types of clinical notes. More recent studies applied Longformer [55], TransformerXL [56], BigBird [57] to clinical coding to process documents of up to 4,096 tokens, but this is still insufficient for the clinical notes. On the other hand, text redundancy (or "Note Bloat" problem [58]) is prevalent in clinical note creation, as measured in recent studies [58-59]. This may impede the performance of deep learning models for code prediction, which may be alleviated through text de-duplication based on text similarity measures [58].

**What are the potential challenges to address for automated clinical coding?**

An empirical fact is that the current BERT-based approaches still do not achieve better performance than CNN-based methods for multi-label classification applied to clinical coding [45,60-61], except for the study [53]. The limitation of BERT may be due to its inefficiency in modelling concept-level information (usually represented in a few keywords or phrases instead of complex relations of tokens in the context) and long documents [60].

Besides, as we stated previously, manual coding is largely based on a standard and implied process with rules applied to the healthcare system, e.g., priority of certain codes, hypothetical mentions, code definitions, mutual exclusion, etc. Future deep learning-based systems need to integrate knowledge reasoning with rules and ontologies to achieve improved and more explainable results.

We list the technical challenges from our work in clinical coding and suggest relevant references below. Some of the challenges are also presented in a different way in a recent, concurrent review in Teng *et al.*, 2022 [14]. The challenges of explainability and few- and zero-shot learning are more relevant to the multi-label classification approach but may be alleviated by the NER+L approach.

- **Creating gold standard coding datasets** – the current widely used benchmark dataset MIMIC-III may have been significantly under-coded [62]. There is a lack of large, openly available, and expert-labelled datasets from Electronic Health Records in this area, and models trained on MIMIC-III may not simply generalise to other datasets due to the difference in length, style, and language (for example, clinical notes in China, Spain, or even the UK). Various expert-labelled coding datasets are also needed for different purposes of using clinical codes (for decision making, diagnosis, epidemiology, etc.), for example, for epidemiology studies to identify deep phenotypes (potentially link to nuanced terminologies like SNOMED CT) from multimodal and multi-source clinical data. Ensuring accurate and publicly available datasets from more healthcare systems for various purposes will better support the clinical NLP community.

- **Coding from heterogeneous, incomplete, and noisy sources** – Clinical coding should be based on *all the relevant documents* of a patient, rather than just discharge summaries as in the majority of recent studies, as discussed in Alonso *et al.*, 2020 [16]. This brings the challenges of long documents as discussed previously. *Structured data*, such as laboratory results, can also be included as a source for coding [49]. *Radiographs* can be useful for coding as well. Besides, real-world data for clinical coders are usually *incomplete* and *noisy*, even for the same type of document (e.g., discharge summary), there is no guarantee that the document is available for all cases and presented in a unified format (i.e. can be hand-written or typed, with various levels of completeness).

- **Explainability of clinical coding** – coders need to understand how the decisions are made by the system. The challenge is more related to the deep learning based multi-label classification approach. Work in this area so far uses label-wise attention mechanisms to highlight key *n*-grams [21], words, and sentences [61,63]. However, the highlighted texts mostly indicate associations instead of causality. Further studies are needed to evaluate the usefulness of highlights for clinical coders and also to integrate more inherently explainable methods, for example, integrating symbolic representations of the coding steps with deep learning.

- **Human-in-the-loop learning with coders' feedback** – to better deploy an automated coding tool into practice, it is essential to involve coders' feedback in the system [10]. The feedback may take different forms, for example, manual corrections, highlights, and rules. The feedback may need to be incorporated into a deep learning system for coding. There may be many rounds of updating the system based on coders' feedback. There were examples in NER+L tools, which are yet to be deployed for clinical coding: in MedCATTrainer

- [64], a dedicated interface is deployed for users to add new concepts, new synonyms and abbreviations, corrections of concepts (of samples selected using active learning), and binary annotations of temporality and phenotyping, then the model is re-run with the feedback; an interface is also designed in SemEHR [65] to allows users to add labels for mentions, which is used to either train a confidence model or to form post-processing rules to refine the results; manually added rules may also be integrated with weak supervision to generate coded data for training [66-67]. A relevant area to human-in-the-loop learning is active learning, with is about selecting the minimum set of most important data for humans to provide annotation feedback; active learning is deployed in NER+L in MedCATTrainer [64], and evaluated in automated coding to potentially reduce human annotations [68].

- **Few-shot and zero-shot learning** - many codes have a low frequency or even no occurrence (or "unseen") in the training data, this is a key problem for multi-label classification with many labels (e.g., 68,000 codes in ICD-10) [38]. The best systems so far to work with low-frequent (<5 times) codes on the MIMIC-III dataset are still below or around 40% recall at $K$ (or the percentage of correct codes in top-$K$ predictions, $K$=10 or 15) [38,47-48]. Better support for few-shot and zero-shot learning will improve the overall coding performance and usage. Knowledge (e.g., descriptions, properties, relations from multiple linked sources, and coding rules) can bridge the gap between the seen and unseen codes, as reviewed in the general domain [69].

- **Adaptation to terminology changes** – how a trained model can be adapted to modified standards for coding or a completely new ontology (for example from ICD-10 to ICD-11 [25])? As we described earlier, ICD-11 is semantically more complex than ICD-10 with a poly-hierarchical backbone structure and the post-coordination of codes. The transition of terminologies may require novel paradigms in deep learning (e.g., self-supervised learning, transfer learning, and meta-learning), accurate ontology matching, concept drift handling, and the above-mentioned robust few-shot and zero-shot learning for new codes with no or few training data.

- **Knowledge representation and reasoning in coding** – finally and most fundamentally, many of the above technical directions suggest to integrate knowledge or semantic information in coding classification systems and ontologies. ICD code descriptions [21,55] and hierarchies [42-43] have been considered in recent studies (and see the blog about hierarchical evaluation [70] for [42]). Other ontologies, such as CCS [71] and code synonyms in UMLS, have been adopted recently to achieve state-of-the-art performance [46,52]. Also, manual coding is mainly based on a standard process and coding guidelines, potentially formalised as a set of rules and terminologies deployed in the healthcare system, for example, the priority of certain codes, the number of codes for each case, the mutual exclusion among certain codes, the rules to code hypothetical cases (e.g., possible and probable), the locally defined specific codes, etc. An example of formalising and integrating rules regarding the mutual exclusion of codes and hypothetical cases with machine learning is presented in the study [30]. These guidelines need to be formally represented in a machine-readable way and to be iteratively integrated into the deep learning based automated coding system.

While multi-label classification is a straightforward formulation of clinical coding, another approach is through named entity extraction and linking or NER+L (for example in the work of MedCAT [36] and the study of rare disease identification [66-67] with SemEHR [65]), although less adopted in the recent literature. NER+L is based on the general approach of clinical information extraction, which is also more recently enhanced by deep learning [72]. NER+L is explainable and feasible, as it inherently links the code to the piece of text in the document and helps handle the long document problem, but the extracted codes still need to be summarised to the final set of codes, and abide by the standard process and guidelines of coding. NER+L methods may help alleviate the coding of few-shot and zero-shot codes by extracting the concepts in the target ontology from clinical notes. A downside of NER+L based coding is that it requires contextual understanding, i.e., the negation, temporality, and experiencer of the extracted concept or code [36,65], which are not needed using the multi-label classification approach. These two formulations (multi-label classification and NER+L) may be combined in the design of a clinical coding system. A recent attempt is to use either text enrichment or multi-task learning to integrate NER+L identified concepts [37], which however does not improve over the multi-label classification approach, and warrants future studies for alternative methods. The study [73] uses NER+L and ontologies to help synthesise clinical notes by replacing words with synonyms or with names of sibling codes (thus to predict the sibling codes) to potentially improve few- and zero-shot coding. Also, the study [62] used NER+L to explore the under-coded problem of clinical coding. The study [74] proposes to rank ICD-10 codes extracted from an off-the-shelf NER+L system for billing code prediction, which better addresses the few- and zero-shot problem than multi-label classification. More benchmarking results for NER+L enhanced methods are needed for comparison.

Automated clinical coding systems also need to be tailored for different purposes (e.g., billing vs. health-related research) and contexts (e.g., countries). For billing purposes, automated coding systems aim at predicting Diagnosis-Related Groups (DRGs) in the US (and Healthcare Resource Groups, HRGs in the UK), which have a smaller number of codes, usually grouped from the full set of ICD codes but can potentially be predicted prior to the ICD coding [75]. For health-related research, automated coding task needs a variety of classification systems (usually with high granularity) for use in case detection or phenotyping, thus other terminologies (e.g., SNOMED CT [76], ORDO [67,77], and ICD-11 in the near future [26]) and customised terminologies (e.g., for sub-stroke phenotyping [78]), and also see the surveys [9,79]. NER+L systems with rule-based inference can help improve the phenotyping when data are scarce to be used for supervised learning [67,80]. Automated coding systems can also be jointly designed with clinical outcome predictions (e.g., readmission and mortality) using deep learning in an end-to-end manner [81]. Also, case detection in some health-related research may favour precision (PPV) than recall (sensitivity) for evaluation [82], which needs to be considered in building and tuning the automated coding system. In terms of other country-related factors, a known issue mentioned earlier in the US is "Note Bloat", where content-importing shortcuts like copy-and-paste are used, which may reduce the time of documentation [58]. The "Note Bloat" phenomenon exacerbates the redundant entry of data in notes that is pulled in or copy-and-pasted from discrete places (e.g., various charts) in the Electronic Health Record (EHR). Training a model to fill codes to the charts needs to remove information from the notes (e.g., ICD codes) that is already present in the charts in the EHR. Also, it is shown that de-duplication of clinical notes improves the performance of prediction tasks, including predicting codes in the DRGs for billing [58]. More country related factors, e.g., billing and insurance, may also affect the system design and would warrant future studies.

Besides, industry organisations, beyond healthcare institutions and academia, play a key role related to automated clinical coding. There are also increasing collaborations between industry and academia. The Epic EHR system is deployed in the University College London Hospital (UCLH) for the management of EHRs. Recently, the CogStack team (including King's College Hospital (KCH), NIHR Mausley Biomedical Research Centre, and UCLH) is collaborating with the UCLH Epic team to integrate an NLP component into the NoteReader interface in the Epic system. The NER+L tool MedCAT is planned to be deployed to populate structured information (by extracting concepts including diagnosis, symptoms, medications, etc.) from newly-created clinical notes to reduce documentation time and verify and complement structured information [83]. Working with five NHS Trusts in England, the CogStack team has also received an AI award from the National Institute of Health Research for developing AI-based clinical coding of medical records (see news from KCH [84]). The project aims to enable more efficient and accurate analysis, free up staff time, and improve research. Industry NER+L APIs (e.g., Amazon Comprehend Medical InferICD10CM [85], Microsoft Text Analytics for health [86] and Google Healthcare Natural Language API [87]) have been released during the last two to three years [88-89] to support clinical concept extraction from texts with price charges. Many technology companies in the industry also provide proprietary solutions and paid services for (semi-)automated clinical coding including Deloitte [90], Optum [91], Capita [92] and CHKS [93]. However, the research access and inner working of the systems are usually not available, leaving it hard to contrast and compare technically. Due to its promising potentials both clinically and financially, the automated coding also attracts great attentions from start-up companies. For example, AKASA in the US is developing a deep-learning based solution, aiming to tackle automated clinical coding adapting a multi-label classification approach. They reported performances with the state-of-the-art results on MIMIC-III full codes, better than human coding in the experiments [51] (also see news [94]). These contribute to the overall picture of the promising potential of automated clinical coding.

**Conclusion**

In this paper, we reviewed the task of automated clinical coding from the perspectives of AI researchers and clinical coding professionals, what it is and why it is an important task, and summarised the challenges of the recent deep learning methods for the task. We then position several key directions for future studies.

While we summarised the *technical* challenges, there are many *organisational* challenges to be addressed to deploy an AI-based coding tool into the clinical coding environment, as reviewed in Campbell & Giadresco [10], where an essential idea is that **coders need to be involved in the model development and deployment stage**. Coders are usually occupied with their coding work and it may not be easy to engage them for system testing. Further research support on projects in medical informatics and computer science is needed to address these challenges.

How far are we from automated clinical coding that is human-centred, explainable, intelligent, and robust to complex real-world scenarios? We cannot give a concrete estimation, but it seems we now have a clearer path and a list of challenges to address. With the growing number of studies and projects in academia and the industry, we look forward to seeing more advances in AI-assisted clinical coding in the next five years and beyond and its application into practice in the near future.

**Data Availability**: The authors declare that all data supporting the findings of this perspective article are available within the paper.

**Acknowledgements**: The work is supported by WellCome Trust iTPA Awards (PIII009, PIII032), Health Data Research UK National Phenomics and Text Analytics Implementation Projects, and the United Kingdom Research and Innovation (grant EP/S02431X/1), UKRI Centre for Doctoral Training in Biomedical AI at the University of Edinburgh, School of Informatics. HD and JC are supported by the Engineering and Physical Sciences Research Council (EP/V050869/1) on "ConCur: Knowledge Base Construction and Curation". HW was supported by Medical Research Council and Health Data Research UK (MR/S004149/1, MR/S004149/2); British Council (UCL-NMU-SEU international collaboration on Artificial Intelligence in Medicine: tackling challenges of low generalizability and health inequality); National Institute for Health Research (NIHR202639); Advanced Care Research Centre at the University of Edinburgh. We thank constructive comments from Murray Bell and Janice Watson in Terminology Service in Public Health Scotland, and information provided by Allison Reid in the coding department in NHS Lothian, Paul Mitchell, Nicola Symmers, and Barry Hewit in Edinburgh Cancer Informatics, and staff in Epic Systems Corporation. Thanks for the suggestions from Dr Emma Davidson regarding clinical research. Thanks to the discussions with Dr Kristiina Rannikmäe regarding the research on clinical coding and with Ruohua Han regarding the social and qualitative aspects of this research. In Figure 1, the icon of "Clinical Coders" was from Freepik in Flaticon, https://www.flaticon.com/free-icon/user_747376; the icon of "Automated Coding System" was from Free Icon Library, https://icon-library.com/png/272370.html.

**Competing Interests:** The authors declare no competing financial or non-financial interests.

**Author Contributions**: HD and HW conceived the research on automated clinical coding. HD, HW, WW, BA, FM discussed the research regularly. HD, HW, BA, WW initiated the writing of this paper. HD and FM discussed the research with practitioners in clinical coding and received their feedback, with support from HW and WW. SJ and JC provided comments on AI for clinical coding as researchers in the field of AI. JM provided comments on the contexts and the design of clinical coding systems from the perspective of the industry. HD drafted the manuscript. All authors edited and revised the manuscript and approved the final version of the manuscript.